# K-Means Segmentation Based-on Lab Color Space for Embryo Detection in Incubated Egg

Shoffan Saifullah [1], Rafał Dreżewski [2], Alin Khaliduzzaman [3], Lean Karlo Tolentino [4], Rabbimov Ilyos [5]

[1] Department of Informatics, Universitas Pembangunan Nasional Veteran Yogyakarta, Jl. Babarsari 2 Yogyakarta, 55281, Indonesia
[2] Institute of Computer Science, AGH University of Science and Technology, Al. Mickiewicza 30, 30-059 Kraków, Poland
[3] Faculty of Agricultural Engineering and Technology, Sylhet Agricultural University, Sylhet-3100, Bangladesh
[4] Department of Electronics Engineering, Technological University of the Philippines, Manila 1000, Philippines
[5] Faculty of Digital Technologies, Samarkand State University, Samarkand, Uzbekistan

**ABSTRACT**

The quality of the hatching process influences the success of the hatch rate besides the inherent egg factors. Eliminating infertile or dead eggs and monitoring embryonic growth are very important factors in efficient hatchery practices. This process aims to sort eggs that only have embryos to remain in the incubator until the end of the hatching process. This process aims to sort eggs with embryos to remain hatched until the end. Maximum checking is done the first week in the hatching period. This study aims to detect the presence of embryos in eggs. Detection of the existence of embryos is processed using segmentation. Egg images are segmented using the K-means algorithm based on Lab color images. The results of the image acquisition are converted into Lab color space images. The results of Lab color space images are processed using K-means for each color. The K-means process uses cluster k=3, where this cluster divides the image into three parts: background, eggs, and yolk. Egg yolks are part of eggs that have embryonic characteristics. This study applies the concept of color in the initial segmentation and grayscale in the final stages. The initial phase results show that the image segmentation results using k-means clustering based on Lab color space provide a grouping of three parts. At the grayscale image processing stage, the results of color image segmentation are processed with grayscaling, image enhancement, and morphology. Thus, it seems clear that the yolk segmented shows the presence of egg embryos. Based on this process and results, the initial stages of the embryo detection process used K-means segmentation based on Lab color space. The evaluation uses MSE and MSSIM, with values of 0.0486 and 0.9979; this can be used as a reference that the results obtained can detect embryos in egg yolk. This protocol could be used in a non-destructive quantitative study on embryos and their morphology in a precision poultry production system in the future.

**Corresponding Author**:

Shoffan Saifullah, Department of Informatics, Universitas Pembangunan Nasional Veteran Yogyakarta, Jl. Babarsari 2 Yogyakarta, 55281, Indonesia
Email: shoffans@upnyk.ac.id

## 1. INTRODUCTION

The checking of the fertility of hatching eggs [1] is a regular practice in hatchery management [2]. About 5-15% of eggs remain infertile in the hatching eggs lot. This checking is done mostly manually using a candling device which is very laborious and time-consuming [3]. The eggs in the hatchery are checked for embryonic development in the first week of incubation (at day 7), i.e., early embryonic stages. If the egg does not have an embryo, it is removed from the incubator to eliminate the contamination. To increase the productivity and efficiency of the poultry industry, various technological approaches are currently addressing towards automation of poultry hatchery practices [4]–[9]. It is very important to develop a protocol for quick and non-destructive detection of an embryo at the early incubation stages. Thus, this study focuses more on embryo detection in the egg image on hatching. Egg images are taken when the eggs have been hatched on the seventh day. It aims to detect the early embryo egg; if the embryo is not seen, the embryo development can be taken and used for consumption.

The process of detection of egg embryos uses the concept of image processing. The image acquisition is based on taking pictures of the egg candling process. The image processing process uses two concepts based on color [10] and grayscale. Color images contain three color components (red, green, and blue), and grayscale only has one color (gray). Color image processing clarifies the embryo detection process by converting it to grayscale. Color image segmentation with the Lab can divide the area of objects according to the detected color uniformity. This segmentation process also applies the K-means clustering algorithm to separate areas that have similarities using distance calculation (Euclidean Distance). In this study, the following process is to clarify the egg embryo by grayscale image processing. The concept implemented is based on improving image quality. Image enhancement is used to clarify the image so that when processed, giving better results. These increased results are used to detect objects by applying morphological operations. Morphological operations can provide a clear picture of an object by eliminating noise.

This research contributes to the development concept of image segmentation based on the Lab color space. Besides, the results of segmentation Lab color images (initial) are processed by grayscaling, image enhancement, and morphology. This article is divided into five parts. The first part explains the background, problems, and state-of-the-art research conducted. The second part describes the relevance and relation of the relationship between this study and previous studies to show the difference in embryo detection. The third part explains the research methods used. The fourth part describes the discussion and results of the experiments carried out. The last part is fifth part is the conclusion part of the topic discussed in this study.

## 2. METHOD
### 2.1. Related Works

This section discusses the relevance and relationship of previous studies related to this research. Topics covered include egg objects, acquisition processes, methods, results, conclusions, and state-of-the-art comparisons of our articles. Based on previous research, identifying egg fertility on hatching machines has been carried out—research on determining the feasibility of eggs with a semi-transmitting mode. Research in [11] uses hyperspectral line-scan imaging. The analysis of this study uses the Principal Component Analysis (PCA) method, and image processing uses a texture co-occurrence method. The results obtained from 88 eggs tested by this method gave 99% accuracy. Eggs that are processed can show significant differences between blood vessels that can be fertilized or not.

CNN classification can distinguish fertile and dead eggs during the ninth-day hatching process [5]. This method is combined with channel weighting (squeeze-and-excitation module) and joint supervision for the classification process. The result is that eggs can be sorted according to existing conditions with a percentage of success of 98.8%. This research is a reference and differentiator from the research we do. Our research is based on segmentation for the detection process of egg fertility.

Prediction of egg fertility can be made with ANN [12]. Egg fertility was predicted using a dataset of 150 chicken eggs. The prediction process with several image processing methods, including preprocessing, color segmentation, and classification algorithms with pattern recognition (ANN). The predictions made have an accuracy of 97% of all processes. The difference with our research is the segmentation method used and the final process to show the presence/absence of an embryo.

Besides, egg image processing related to embryos based on thermal imaging-based images [13]–[16], hyperspectral imaging [17], [18], and digital cameras with the concept of candling [19] has been extensively studied. This research uses image processing with a camera using the concept of candling. Thus, the related research base used as a reference is the identification/detection of egg embryos with digital cameras or the like.

This research refers to previous studies. Associated with image processing in egg image processing associated with egg embryos [20]–[23]. This study uses the concept of segmentation with K-means segmentation based on Lab color space. Besides, embryo detection is based on grayscale images by image processing using image enhancement and morphology.

### 2.2. Research Method

This section explains the methods and stages associated with the segmentation process with color imagery and continues with its grayscaling (Fig. 1) [24]. The method is divided into two main parts to get segmentation results. The first part is to process images based on color images using Lab color space to get segmented color image results. In the second part, this stage is a continuation of the first stage carried out to clarify the object of the embryo in grayscale egg images. This last stage produces images of embryos detected to have image clarity. The method used uses image enhancement and morphology.

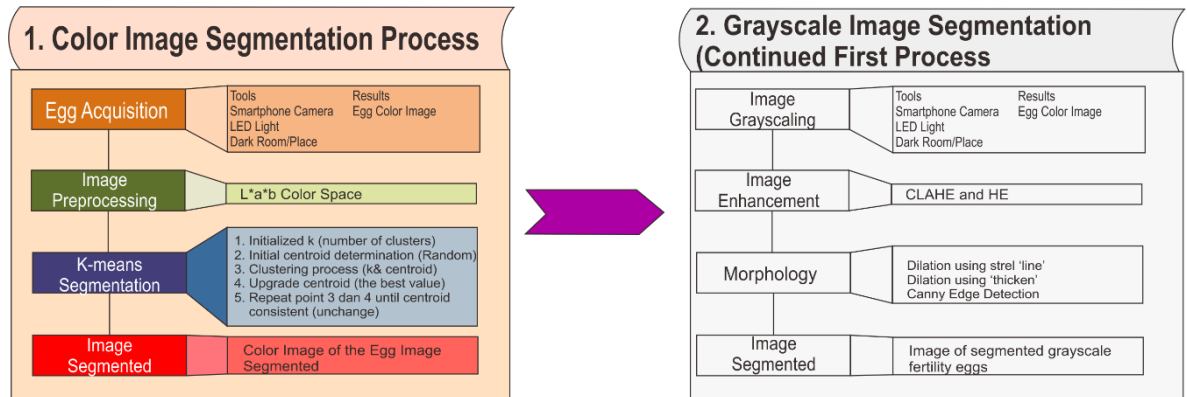

**Fig. 1.** The image segmentation process is based on (a) Lab colors and sequences based on (b) grayscale image conversion

### 2.2.1. Tools and Process of Image Acquisition

Image acquisition is the primary initial stage of image processing. The aim is to digitize objects into space that a computer can process (Fig. 2(b)). The acquisition in this article, as has been done in previous studies [20], [21], [25]. Tools used in the study include smartphone cameras and LED lights (Fig. 2(a)). The acquisition process requires a dark place/room condition to be processed at night or in a room without light.

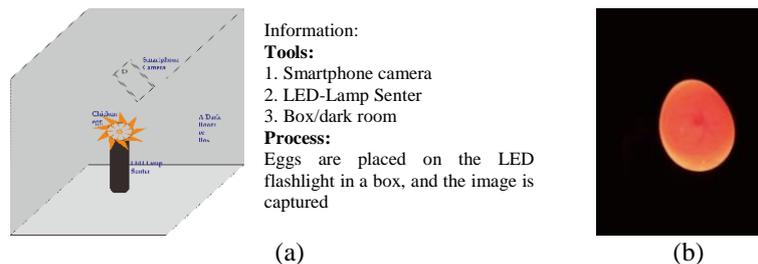

**Fig. 2.** (a) the design system and (b) the result of the image acquisition process

### 2.2.2. Lab Color Space on Image Preprocessing

Image preprocessing in the initial stages uses the conversion of color images (RGB) into Lab color images. Lab color images use the concept of coloring based on luminance or lightness (L) of white-black and chromatic color components (a. red-green and b. yellow-blue) [26]. The color space lab is a development of CIEXYZ color space [27], with L values represented in the values 0 (black) and 100 (white). The range of the a* and b* coordinates does not depend on the converted color space, as it uses the X and Y of RGB. As for the chromatic color representation (red-green and yellow-blue), each is between -128 to 128. Labs have the concept that colors should not be based on the device displayed (device-autonomous).

The Lab Model is a simple color model, where time is an absolute criterion that compares all color models. Changing the Lab color model from RGB requires the conversion of CIEXYZ to Lab. The conversion process requires the calculation process as in equation (1)-(7) [27], [28] with a default of σ = 6/29.

$$L^* = 116 g\left(\frac{Y}{Y_n}\right) - 16 \tag{1}$$

$$a^* = 500 \left(g\left(\frac{X1}{X1_n}\right) - g\left(\frac{Y1}{Y1_n}\right)\right) \tag{2}$$

$$b^* = 200 \left(g\left(\frac{Y1}{Y1_n}\right) - g\left(\frac{Z1}{Z1_n}\right)\right) \tag{3}$$

$$f(t) = \begin{cases} \sqrt[3]{t}, & if\ t > \sigma^3 \\ \dfrac{t}{3\sigma^2} + \dfrac{4}{29}, & otherwise \end{cases} \quad (4)$$

$$X1 = X1_n g^{-1}\left(\dfrac{L^* + 16}{116} - \dfrac{a^*}{500}\right) \quad (5)$$

$$Y1 = Y1_n g^{-1}\left(\dfrac{L^* + 16}{116}\right) \quad (6)$$

$$Z1 = Z1_n g^{-1}\left(\dfrac{L^* + 16}{116} + \dfrac{b}{500}\right) \quad (7)$$

where, $L^*$ coordinate ranges from 0 to 100. The $X, Y, Z$ are the color stimulus that is considered. $X_n, Y_n, Z_n$ are the specified white reference illumination. $f(t)$ is a domain division function into two parts.

### 2.2.3. K-Means Segmentation Process

The segmentation process in this study uses the concept of clustering. The image is divided into three parts: background, eggs, and yolk eggs [29], [30]. This segmentation process uses the K-means algorithm with a value of $K = 3$.

The steps in the k-means process [31], [32] are shown as follows:
1. Determining the K value used in the clustering process
2. Randomly selecting K centroid points for initial determination
3. The process of grouping data by forming K clusters used the centroid point of each cluster, which was selected earlier. This process calculates the distance in (8), namely Euclidean Distance.

$$d(i,j) = \sqrt{\sum_{k=1}^{p}(x_{ik} - x_{jk})^2} \quad (8)$$

Euclidean distance ($d(i,j)$) is the object's distance between $i$ and $j$. The calculation process uses the p-value as a dimension of data that illustrates the number of values to be processed. The Euclidean distance calculation will reduce object $i$ ($x_{ik}$) coordinates with object $j$ ($x_{jk}$) on the same dimension $k$. The centroid point renewal process is calculated by (9).

$$\mu_k = \dfrac{1}{N}\sum_{q=1}^{N_k} x_q \quad (9)$$

The calculation of centroid points in the k-cluster ($\mu_k$) is processed by adding all the data to the $q$ ($x_q$) divided by the amount of data in the k-cluster ($N_k$). The condition used is to use the value $q = 1$ up to many data.
4. Updating the centroid point value to get the best value
5. Repeating processes 3 and 4 until the value of the centroid point is consistent/stable (unchanged)

### 2.2.4. Image Grayscaling and Enhancement

This study uses grayscaling to convert the image results of color segmentation into grayscale images. The conversion process uses Eq. (10). Each color component (Red, Green, and Blue) is used to calculate the average to a grayscale image. The mean of these images can be calculated with the same value or different according to the color composition. Red is multiplied by 0.2989, green is multiplied by 0.587, and blue is multiplied by 0.1141.

$$G' = 0.2989 * R + 0.587 * G + 0.1141 * B \quad (10)$$

Image enhancement improves the image [33] before processing [34]. Image enhancement in this study uses a combination of Histogram Equalization (HE) and Contrast Limited Histogram Equalization (CLAHE) [35]. The image will be improved based on its histogram value [36]. The histogram is processed using grayscale images.

The HE process is used because of the spread of intensity of pixels in uneven images, such as too dark/bright images. Repair HE [37] is processed using Eq. (11).

$$H(X_k) = \frac{n_k}{W}, where\ 0 \leq k \leq L-1 \tag{11}$$

HE is processing in grayscale images using (11), where $L$ is the degree of gray, $n_k$ is the number of pixels with gray degrees $k$ (dynamic range $\in [X_0, X_{L-1}]$), and $W$ is the number of all pixels. CLAHE is the development of HE whose histogram has a limit value [38] (with a maximum height limit of the histogram). CLAHE is calculated with a histogram boundary clip limit Eq. (12).

$$\beta = \frac{M}{n}\left(1 + \frac{\alpha}{100}(s_{max} - 1)\right) \tag{12}$$

Based on Eq. (12), the value of $M$ is the size of the region size. $N$ is the value of grayscale, and $a$ is a clip factor as an addition to the limit of the histogram, which is range 0-100.

### 2.2.5. Morphological Method

Image morphology aims to change the shape of the original image and improve segmentation results. This method is processed using a two-image approach, namely, binary and grayscale images. Morphological processes are available in many ways, but this research uses dilation, morphology 'thicken,' edge detection.

Dilation is a morphology to enlarge the size of the object segments by adding layers around the object (thickening of objects in the image). This thickening uses the form of structuring element (*strel*) lines or lines. In this study, the dilation used is the 'line' *strel* with a length of 1 and 45 degrees. The dilation process is processed by surgery (Eq. (13)).

$$A \oplus B \tag{13}$$

In addition to the thickening, the following process is thickening. Thickening is used to thicken the selected area of the foreground pixels in the binary image. Its application can be made by determining the estimated convex hull (shape) and determining the object's skeleton Eq. (14)-(15).

$$A \odot B = A \cup (A \circledast B) \tag{14}$$

$$A \odot \{B\} = ((\ldots((A \odot B^1) \odot B^2)\ldots) \odot B^n \tag{15}$$

In edge detection, the processed image is similar with a different brightness level. Detection results in the form of edges of objects detected in the image. Thus, objects can be marked to improve the image details due to imperfect acquisition processes (for example, motion blur). In this study, we used a canny operator for optimal edge detection with Gaussian Derivative Kernel. So the results given show a smooth edge detection through noise filtering from the initial image. In the edge detection process using the canny operator, there are several processes, including grayscaling, Gaussian filter, intensity gradients, Non-Maximum Suppression, Double Thresholding, Edge Tracking by Hysteresis, and Cleaning Up. The process is done using Matlab with code as in Eq. (16).

$$Ed\_canny = edge(black\_white\_image, 'canny'); \tag{16}$$

Edge detection produces margins that detect objects. In this study, segmentation in the yolk section, embryos were identified in it. The results show that the object is visible with the detection.

### 2.2.6. Evaluation Using SSIM and MSE

In the final evaluation of this study, it is necessary to check the structural similarity index method (SSIM). SSIM is a perception-based method concerning image degradation on changes in perception in structural information [39]. SSIM processes calculations based on luminance, contrast, and structure. The SSIM index is calculated on various image windows. The size between two $x$ and $y$ windows with the available size N × N. The SSIM formula can be seen in Eq. (17).

$$SSIM(x,y) = [l(x,y)]^\alpha [c(x,y)]^\beta [s(x,y)]^\gamma \tag{17}$$

In addition to SSIM, evaluations are carried out using MSE (Mean Squared Error), and the calculation is shown in Eq. (18).

$$MSE = \frac{1}{mxn}\sum_{i=0}^{n-1}\sum_{j=0}^{m-1}[f(i,j)-g(i,j)]^2 \qquad (18)$$

## 3. RESULTS AND DISCUSSION

The section described the implementation of methods and analysis related to segmentation with color images and continued with its gray scaling. This article is divided into two main parts to get segmentation results. The first part included processing the image based on the image's color using the Lab color space to get segmented color image results. The second part, this stage, is a continuation of the first step carried out to clarify the object of the embryo in grayscale egg images. In this last stage, the result of the research produces images of embryos that are detected with clear images.

### 3.1. Image Preprocessing Results

Based on the image acquisition, the initial stage aims to digitize objects into space that a computer can process. Smartphone cameras and LED lights to capture images in a dark place/room condition. This process should be done at night or in a room without light. The image acquisition results in this study were in the form of color images (Fig. 3). The resulting image size is 892×1191, as shown in Fig. 3(a). The original image is processed by cropping to get the image with the whole egg object (Fig. 3(b)). This cropping aims to eliminate background objects that are not needed.

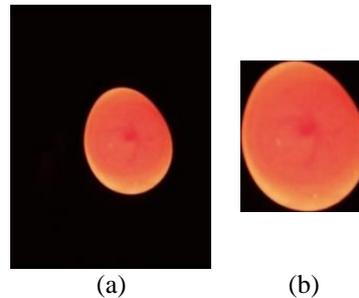

(a)        (b)

**Fig. 3.** The initial image of this research used (a) image acquisition results and (b) an image cropped from the original image

Cropping image becomes a source for further image processing, where this image has a size of 582×778. The object than the processed mind more highlights this part. The image in Fig. 3(b) is an image of an egg detected by an embryo. The embryo egg image [40], [41] has a characteristic in the egg circle, and a network looks faint. The following process will clarify the picture so that the components of the embryo are visible.

### 3.2. K-means Segmented Image

Image preprocessing in the initial stages uses the conversion of color images (RGB) into Lab color images. This conversion is to divide the domain according to the factors of the color space Lab. The L domain in the Lab can be shown in Fig. 4(a) shows the image's luminance. In contrast, fields a (Fig. 4(b)) and b (Fig. 4(c)) show their particular combinations in red-green and yellow-blue sequences.

Based on the image of the acquisition results, the results have been cropped according to the egg object (Fig. 3(b)). Lab results (Fig. 4) can be combined into one image (Fig. 5(b)).

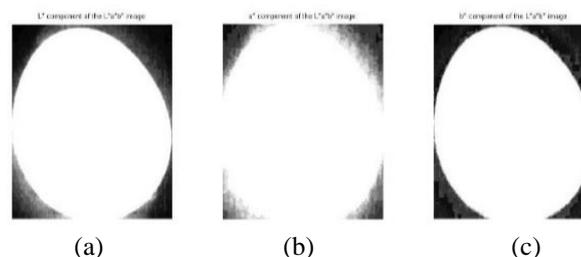

(a)                  (b)                  (c)

**Fig. 4.** RGB to Lab conversion image for each component (a) L (luminance), (b) a (red-green), and (c) c (yellow-blue)

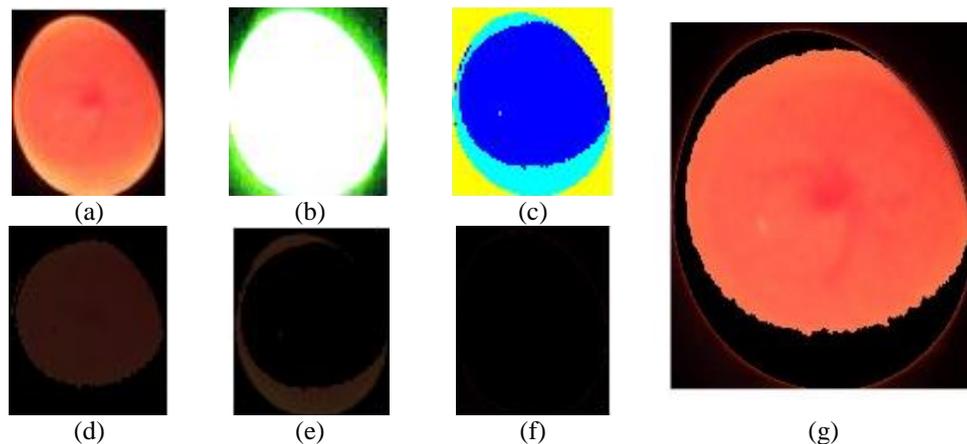

**Fig. 5.** The results of the processes of the first stages of the image are (a) the initial image (the results of cropping), (b) the process of image conversion with Lab color space, (c) clustering with K-means, (d) the first cluster, (e) the second cluster, and the third cluster, and (g) the result of segmented color images

Based on Fig. 5, the result of the Lab color space process got the image segmented. Segmentation of the embryo detection and edge detection of the egg. The last method of detection is cropped and converted to image grayscale. The grayscale result was processed by image enhancement and morphological process.

### 3.3. Grayscaling of Image Segmented Result

This stage produces image segmentation results (Fig. 6(a)) based on color are processed by cropping (Fig. 6(b)) and grayscaling (Fig. 6(c)). This process aims to change segregation's color image, resulting in grayscale images.

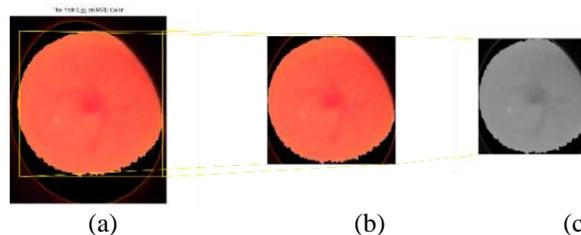

**Fig. 6.** The resulting image is (a) color segmentation, (b) processed by cropping, and (c) converted to grayscale

The grayscale image provided is used for image enhancement and morphology of the image at a later stage. The grayscale image in Fig. 6(c) is obtained by dividing each color (RGB) into one color (grayscale) according to (1).

### 3.4. Image Enhancement and Morphological Process for Embryo Egg Detection

Grayscaling image results are grayscale images. Grayscale images are processed using image enhancement. Image enhancement in this study applies a combination of image improvement based on the histogram. The method used in image improvement is HE and CLAHE. The results of the HE process are shown in Fig. 7(a). Then it is processed again with the repair using the CLAHE method (Fig. 7(b)). The results of this quality improvement give a picture of egg yolk-containing embryos.

The image enhancement becomes a reference for morphological processing. The result of the image enhancement is processed in dilation with a combination of 'line' and 'thicken' strings (Fig. 7(c-d)). The 'line' string provides more apparent imaging results than previous imagery. This processing requires a grayscale image to be converted into a black and white image (BW).

BW images are processed according to processes (4)-(6) to obtain image results, such as Fig. 7(e) for the image improvement process with Histogram Equalization. Contrast the difference in the initial grayscale image with the BW results with HE obtained by more precise image characteristics. The object shown is the embryo from the egg image. The second improvement process used CLAHE to get a smoother and brighter appearance.

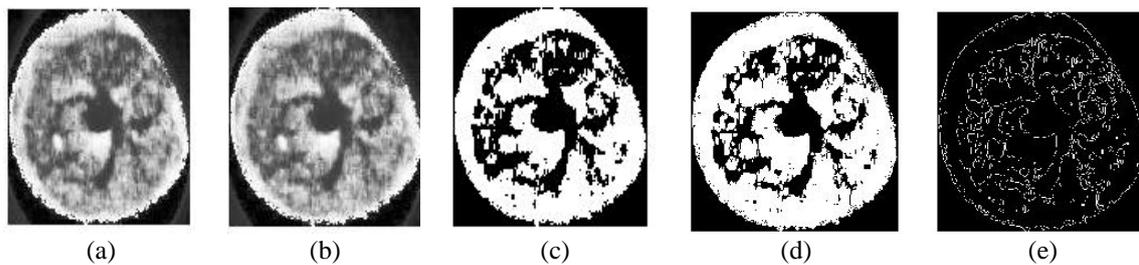

(a)　　　　　　(b)　　　　　　(c)　　　　　　(d)　　　　　　(e)

**Fig. 7.** The results of image processing with image enhancement with the methods (a) HE, (b) CLAHE, and (c), (d) Morphological methods with dilation with the 'line' and 'thicken' strings, and (e) edge detection using the canny method

### 3.5. Image Evaluation using MSE and MSSIM

Image testing was carried out using the Structural Similarity Index Measure (SSIM) based on the research data. The SSIM method shows that embryonic image data testing results in SSIM values close to 1, with an average value of 0.9979.

The SSIM calculation in an embryo egg image has a value distribution between 0.9257 to 1. Thus, based on the SSIM distribution chart, the average value of the SSIM (MSSIM) produced is 0.9979. This number shows that the resulting value gives excellent results where the resulting value is close to value 1. Besides, this result also indicates that the similarity value generated can be used for subsequent processes because the information conveyed does not have much difference. A sample of this SSIM calculation can be shown in Fig. 8. This calculation is carried out by many research data, namely 200 image data with the categories of fertile and infertile each 100 data.

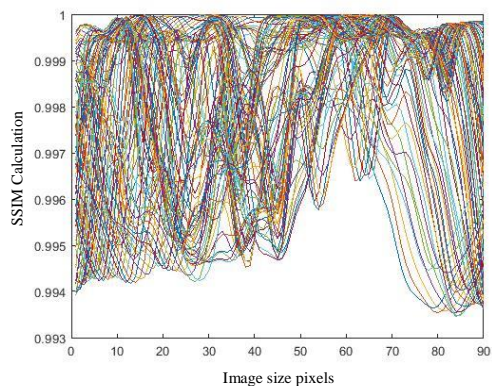

**Fig. 8.** A sample of the SSIM graph calculation by Matlab on an image of chicken egg embryo detection

Besides, the evaluation uses MSE to check errors based on the output image from the initial image based on the BW image. MSE produced is equal to 0.0486 for all data that has been calculated from each data. These results indicate that the resulting image still has similarities with the initial image. It can be used as a reference for the segmentation process for detecting embryos in eggs.

### 4. CONCLUSION

This research revealed that K-means Segmentation with Lab color space could group images into several parts. The image of the processed egg is divided into three parts, namely, background, egg, and yolk. Thus, the color space Lab process with k-means can computerize image objects according to the criteria of image uniformity differences. The result can be clarified with the second stage using grayscale image-based processing methods so that the embryo in the egg yolk image can be seen clearly. The evaluation uses MSE and MSSIM, with values of 0.0486 and 0.9979. This method can be used for real-time application in precision hatchery practices in the egg and poultry industry.

### Acknowledgments

The authors expressed their gratitude to UPN "Veteran" Yogyakarta, especially the Department of Information Engineering and LPPM, who assisted in making this article for publication. We also thank the Institute of Computer Science, AGH University of Science and Technology, who supported this publication.

## BIOGRAPHY OF AUTHORS

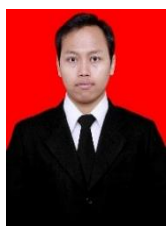

**Shoffan Saifullah** received the Bachelor's Degree in Informatics Engineering from Universitas Teknologi Yogyakarta, Indonesia, in 2015, and Master's Degree in Computer Science from Universitas Ahmad Dahlan, Yogyakarta, Indonesia, in 2018. He is currently a lecturer in Universitas Pembangunan Nasional "Veteran" Yogyakarta Indonesia. His research interests include image processing, computer vision, and artificial intelligence. Email: shoffans@upnyk.ac.id

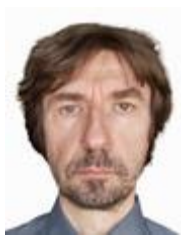

**Rafał Dreżewski** received the M.S., Ph.D., and D.Sc. (Habilitation) degrees in computer science from the AGH University of Science and Technology, Cracow, Poland in 1998, 2005, and 2019, respectively. Since 2019, he has been an Associate Professor with the Institute of Computer Science, AGH University of Science and Technology, Cracow, Poland. He is the author of more than 70 papers. His research interests include bio-inspired artificial intelligence algorithms and agent-based modeling and simulation of complex and emergent phenomena. Email: drezew@agh.edu.pl

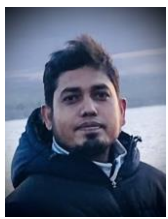

**Alin Khaliduzzaman** is a faculty member at Sylhet Agricultural University, Bangladesh. He obtained a Master's and Ph.D. in Bio-Sensing Engineering from Kyoto University, Japan. He has also served as a JSPS postdoctoral fellow in Japan. During his Ph.D. studies, he opened up a new research domain in the scientific world in the field of non-destructive chicken egg research and pioneered the embryo grading concept for precision hatchery practices. He has also been awarded several conferences and Young Researcher's Awards due to his non-destructive egg and poultry research by various academic societies. Email: khaliduzzamanfetsau2014@gmail.com

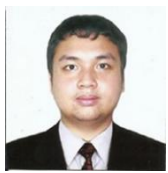

**Lean Karlo S.** Tolentino received a B.S. degree in electronics and communications engineering from the Technological University of the Philippines (TUP), Manila, Philippines, in 2010, and an M.S. degree in electronics engineering major in microelectronics from Mapúa University, Manila, in 2015. He is currently working toward a Ph.D. degree in electrical engineering at the National Sun Yat-sen University, Kaohsiung, Taiwan. Since 2015, he has been an Assistant Professor with TUP. From 2017 to 2019, he was the Head (Chair) of the Department of Electronics Engineering, TUP. He was designated as the Director of the TUP's University Extension Services Office (UES), Manila, from 2019 to 2020. His research interests include artificial intelligence, IC design, and power electronics. Email: leankarlo_tolentino@tup.edu.ph

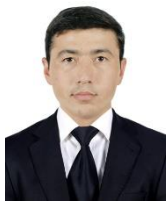

**Ilyos Rabbimov** received a Bachelor's degree in applied mathematics and informatics from Samarkand State University, Uzbekistan, in 2013, and a Master's degree in applied mathematics and information technologies from the Samarkand State University, Uzbekistan, in 2015. He is currently a lecturer at Samarkand State University, Uzbekistan. Furthermore, he is a Visiting Research Fellow with the University of Hertfordshire, where he focuses on artificial intelligence, natural language processing, and image processing. His research interests include image processing, sentiment analysis, and artificial intelligence. Email: ilyos.rabbimov91@gmail.com